\documentclass[journal]{IEEEtran}
\usepackage[T1]{fontenc}
\usepackage{booktabs}
\usepackage{graphicx}
\usepackage{comment}
\usepackage{amsmath,amssymb} 
\usepackage{color}
\usepackage{epsfig}
\usepackage{xcolor}
\usepackage{colortbl}
\usepackage{multirow}
\usepackage{enumitem}
\usepackage{bigstrut}
\usepackage{tabularx}
\usepackage{tabulary}
\usepackage{hhline}
\usepackage[pagebackref=true,breaklinks=true,colorlinks]{hyperref}
\usepackage{cite}
\usepackage{amssymb}
\usepackage[export]{adjustbox}
\definecolor{Gray}{gray}{0.9}

\newlength{\nameWidth} 
\newlength{\eqWidth} 
\newlength{\paraWidth} 
\newcommand{\tableEquation}[6]{ 
	\parbox{\nameWidth}{\centering #2 \cite{#6}} &  
	\parbox{\eqWidth}{\centering
		\begin{equation} \label{eq:#3}
		#4
		\end{equation}
	} 
}

\ifCLASSINFOpdf
\else
\fi
\ifCLASSOPTIONcompsoc
  \usepackage[caption=false,font=normalsize,labelfont=sf,textfont=sf]{subfig}
\else
  \usepackage[caption=false,font=footnotesize]{subfig}
\fi
\usepackage[switch]{lineno}

\begin{document}
\nolinenumbers
\renewcommand{\linenumberfont}{\normalfont\bfseries\small\color{blue}}
\linenumbersep 15pt\relax

%
\title{Image-Based Stability Quantification}
%
%
%

\author{Jesse~Scott,~\IEEEmembership{Member,~IEEE,}
        John~Challis,\\
        Robert~T.~Collins,~\IEEEmembership{Senior Member,~IEEE,}
        and~Yanxi~Liu,~\IEEEmembership{Senior Member,~IEEE}

\thanks{J.~Scott, R.T.~Collins and Y.~Liu, School of Electrical Engineering and Computer Science, Penn State University, University Park, PA, 16802 USA \\
e-mail: 
\{jus121, rtc12, yul11\}@psu.edu }
\thanks{J.~Challis, Biomechanics Laboratory, Kinesiology Dept., 
Penn State University, University Park, PA, 16802 USA   
e-mail: jhc10@psu.edu}
\thanks{Manuscript submitted July 12, 2022; Revised Oct 20, 2022}
}

\maketitle

\begin{abstract}
Quantitative evaluation of human stability using foot pressure/force measurement hardware and motion capture (mocap) technology is expensive, time consuming, and restricted to the laboratory. We propose a novel image-based method to estimate three key components for stability computation: Center of Mass (CoM), Base of Support (BoS), and Center of Pressure (CoP). Furthermore, we quantitatively validate our image-based methods for computing two classic stability measures, CoMtoCoP and CoMtoBoS distances, against values generated directly from laboratory-based sensor output (ground truth) using a publicly available, multi-modality (mocap, foot pressure, two-view videos), ten-subject human motion dataset. Using Leave One Subject Out (LOSO) cross-validation, experimental results show: 
1) our image-based CoM estimation method (CoMNet) consistently outperforms state-of-the-art inertial sensor-based CoM estimation techniques;
2) stability computed by our image-based method combined with insole foot pressure sensor data produces consistent, strong, and statistically significant correlation with ground truth stability measures (CoMtoCoP r~=~0.79 p~<~0.001, CoMtoBoS r~=~0.75 p~<~0.001);
3) our fully image-based estimation of stability  produces consistent, positive, and statistically significant correlation on the two stability metrics (CoMtoCoP r~=~0.31 p~<~0.001, CoMtoBoS r~=~0.22 p~<~0.043).
Our study provides promising quantitative evidence for the feasibility of image-based stability evaluation in natural environments.
\end{abstract}


\vspace{-5pt}
\section{Introduction}
\begin{figure}[!b] \centering
    \vspace{-15pt}
	\includegraphics[width=.76\linewidth]{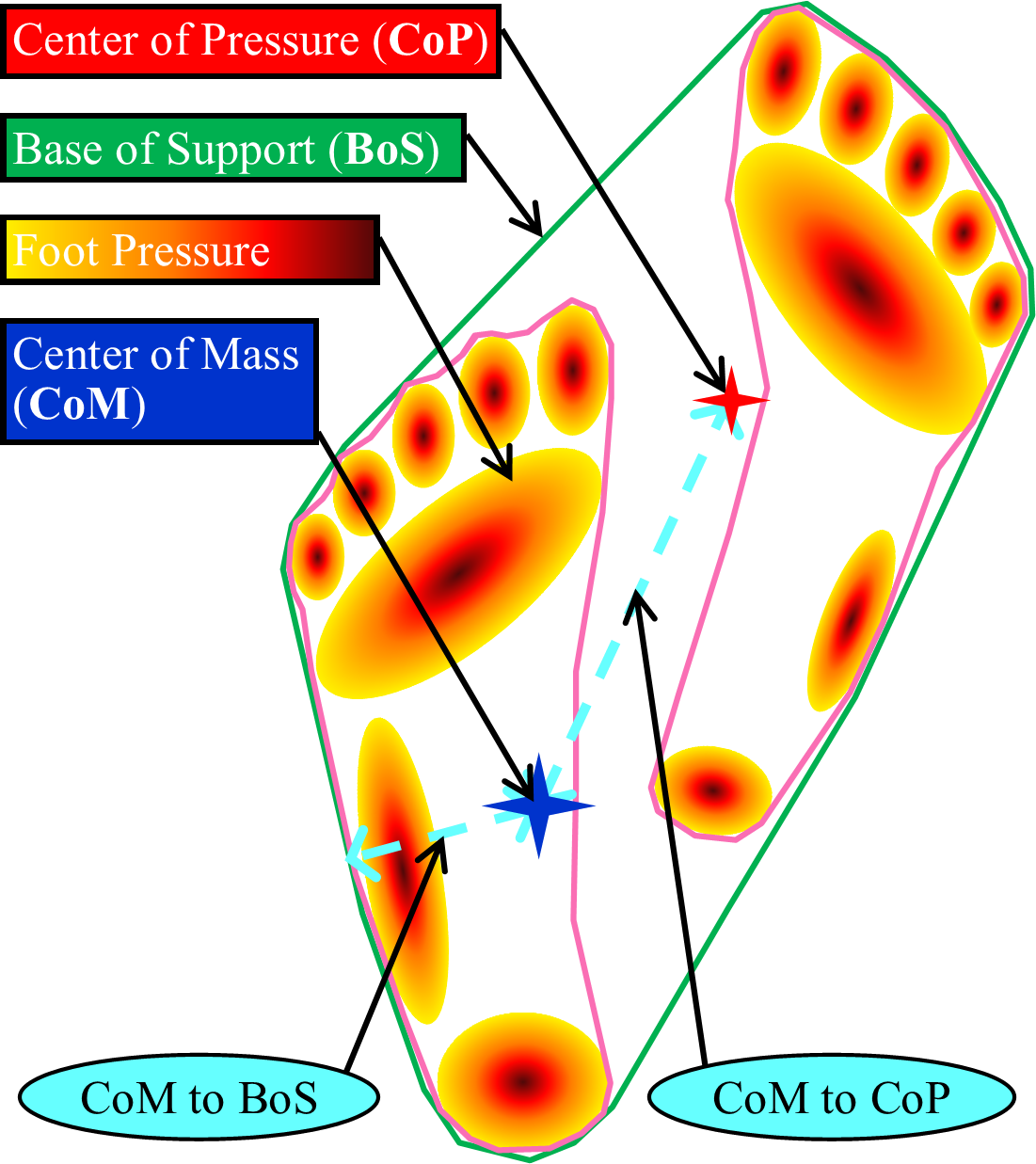}
	\vspace{-10pt}
	\caption{Stability components and two stability metrics (CoMtoCoP and CoMtoBoS) relative to localized foot pressure with CoP (red star), CoM (blue star),
	BoS (green border), foot pressure (yellow/red/brown gradations), and stability metrics (cyan lines)~\cite{scott_2020}.}
	\label{fig:stability_components}
\end{figure}

\IEEEPARstart{F}{alls} in the elderly are an important worldwide health problem~\cite{englander_1996}, and their frequency increases with age~\cite{geriatrics_2001}. Therefore, frequent and accurate monitoring of human motion stability, especially for the elderly, becomes more and more necessary~\cite{parkkari_etal_1999,sterling2001geriatric,CDCStats_2015}. Three essential and commonly used component measures for human stability assessment are: Base of Support (BoS), Center of Pressure (CoP), and Center of Mass (CoM)~\cite{winter_1995, Hof_2005} (Fig.~\ref{fig:stability_components}). Accurate estimation of these key components is currently expensive and time-consuming, involving  foot pressure/force plates,  motion capture hardware/software, and tedious post-processing of error-prone sensor data~\cite{Challis2001forceplate,Whittle2007}. For these reasons, stability measurement is usually restricted to a laboratory environment. A fully or partially image-based method for stability monitoring would be an attractive alternative for deployment in rehabilitation or elder care facilities where unencumbered long-term monitoring could have significant clinical value and allow for preventative or timely corrective interventions to reduce falls.

In recent years, human pose extraction from images and video has become an active research area in computer vision and machine learning~\cite{zheng2020deep}. However, little work has been done in image-based mapping of body kinematics (pose) to dynamics (foot pressure/force). In an initial study~\cite{scott_2020}, we demonstrated the feasibility of predicting \emph{foot pressure} from images of human pose. We take a step further in this work to explore the feasibility of predicting BoS, CoP, and CoM measures from visual input, and of using these image-based estimates to compute two classic human stability metrics: CoMtoCoP and CoMtoBoS (Fig.~\ref{fig:stability_components}, Table~\ref{tbl:stability_metrics}).

Using Taiji (a.k.a.~Tai Chi) performance data provided by the  PSU-TMM100 dataset~\cite{scott_2020}, this work makes the following main contributions:\\
1) developing and validating an image-based machine learning algorithm for CoM estimation from image data;\\
2) assessing two stability metrics (Table~\ref{tbl:stability_metrics}) with a thorough comparison  using component values CoP, BoS, and CoM obtained from either image-based or sensor-based (ground truth) measurements ($2^3$~=~8 combinations evaluated);\\
3) finding that a fully image-based approach (eliminating the need for foot pressure sensors and motion capture) produces stability estimates that are positively correlated with ground truth (CoMtoCoP r~=~0.31 p~<~0.001, CoMtoBoS r~=~0.22 p~<~0.043); and\\
4) finding that insole foot pressure data combined with image-based foot localization and CoM prediction (eliminating need for motion capture hardware) produces stability estimates that are strongly correlated with ground truth estimates (CoMtoCoP r~=~0.79 p~<~0.001, CoMtoBoS r~=~0.75 p~<~0.001).

The paper is organized as follows: Section~\ref{Background} covers background information on stability components and metrics, image-based estimation of dynamics, and the PSU-TMM100 dataset that provides ground truth sensor measurements in this research. Section~\ref{computing_stability} covers calculation of the stability components from ground truth data and image-based data, while Section~\ref{selected_metrics} covers the stability metric calculations. Section~\ref{results} quantifies and visualizes image-based estimates for CoM, CoP, BoS, CoMtoCoP, and CoMtoBoS, and compares them with sensor-based ground truth estimates. Section~\ref{Conclusion} summarizes the results.

\begin{table}[!t] \centering
	\caption{Selected biomechanical stability metrics determined from $CoM$, $CoP$, and $BoS$.}\vspace{-5pt}
	\begin{tabular}{cc}\toprule
		Name & Equation  \\ \midrule
		\tableEquation{1}{CoMtoCoP}{CoMtoCoP}{\lVert CoM - CoP \rVert_2}{The $\ell_2$ distance between the CoM and the CoP.}{jian_1993,Chaudhry_2011} \\ \cmidrule(l){1-2}
		\tableEquation{2}{CoMtoBoS}{CoMtoBoS}{ \begin{split} & \qquad\enspace\lVert CoM - BoS_{nearest} \rVert_2 \\ &  \begin{cases} 
				positive  & \text{if } CoM \text{ is inside BoS}\\
				negative  & \text{if } CoM \text{ is outside BoS}
		\end{cases}\end{split}}{The shortest distance between the BoS border and the center of mass (CoM).}{Lugade_2011}
		\vspace{-5pt} \\ 
		\bottomrule
	\end{tabular} \vspace{-15pt}
	\label{tbl:stability_metrics} 
\end{table}

\vspace{-10pt}
\section{Background} \label{Background}
In a review of video-based measurement for human movement science, Seethapathi~\emph{et~al.}~\cite{Seethapathi2019} indicate that improving kinematic accuracy and estimating dynamics (contact forces) are the key to  practical use of computer vision as a tool in biomechanics. Upright human body stability is often investigated by examining relative motion of the CoM compared to the BoS or CoP, which requires measurement of pose and contact forces. Currently, no research exists that uses standard RGB video cameras to automatically determine human body stability during complex~\emph{actions}. Our approach is novel in being the first to use pose and ground force dynamics computed solely from video for stability analysis.

    \subsection{Balance and Stability}
    Balance and stability are terms often used interchangeably to describe how well an individual is able to keep from falling. In kinesiology, \textbf{balance} describes maintaining static position without significant movement; e.g., balancing on one foot~\cite{winter_1995}. \textbf{Stability} describes continuing dynamic movement of the body while preventing an uncontrolled fall or unplanned movement~\cite{murray_1967}.
    Humans have a natural physiological ability to sense their own balance and maintain stability~\cite{Asslander_2015} but there is a difference between perception and physical ability that is not easily determined~\cite{Winter2009}. Computational evaluation of quantified stability uses specialized equipment like force plates to capture 3D foot forces and motion capture technology to measure body movements~\cite{winter_1995},  constraining research to a laboratory setting and limiting its ecological validity.

        \subsection{Quantification and Metrics}\label{chebel}
        A comprehensive review by Bruijn~\emph{et~al.}~\cite{Bruijn_2013} breaks stability metrics into three categories: i) ability to recover from small perturbations, derived from dynamical systems theory and biomechanics, ii) ability to recover from larger perturbations, and iii) determining the maximum controllable perturbation. 
        
        The size of the BoS determines the tolerable condition during gait termination~\cite{Pai_1997} and unexpected perturbation recovery in upright stance~\cite{hof_2016}. King~\emph{et~al.} identify a decrease in the size of the functional BoS with increasing age~\cite{king_1994}. Given that the BoS is a determinant of upright stance balance and gait stability, its quantification is an important feature during the analysis of human movement. BoS boundaries are established in~\cite{haibach_2006} by subjects swaying in a circular fashion, defining the boundary by the maximum CoP positions. Force plate and motion analysis data are used to determine a BoS of subjects walking in~\cite{suptitz_2013}, but these testing conditions limit data collection to a laboratory. Body segment inertial properties and motion analysis data are used in~\cite{pillet_2010} to generate estimates of the CoP motion during gait, while CoP motion is determined for sidestep movements by exploiting convolutional neural network (CNN) models in~\cite{johnson_2019}. 
        
        Previous work has reported using the center of the hip joints as an approximation for CoM, e.g. \cite{Ng_2020}. More recently, Chebel~\textit{et al.}~\cite{chebel_2021} present a state-of-the-art neural network for 3D CoM estimation using two subject height measurements (head and hip) and 11 inertial sensors measuring joint angles as input while subjects either squat in place or walk.  
        They report RMSE errors in a componentwise format that works out to total 3D mean error of 18.1~mm for a full body model tested on new subjects.
        In comparison, our CoMNet (Section~\ref{CoM_prediction}), a neural network predicting 3D CoM trained on image-based poses only and tested on unseen subjects, has a mean error of 17.6~mm.

\begin{figure*}[!t] \centering
\setlength{\tabcolsep}{1pt}
    \subfloat[]{\centering \includegraphics[width=0.435\linewidth,valign=m]{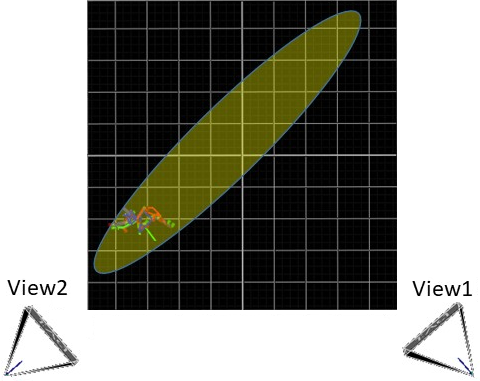} \label{fig:PSU_TMM100:a}}
    \hfill
    \subfloat[]{\centering
        \begin{tabular}{cr}
    	\multicolumn{2}{c}{\includegraphics[width=.29\linewidth,valign=m]{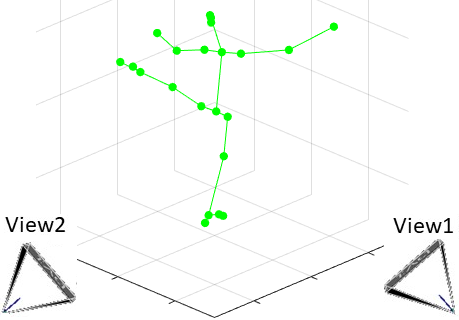}} \\
    	\includegraphics[width=.12\linewidth,valign=m]{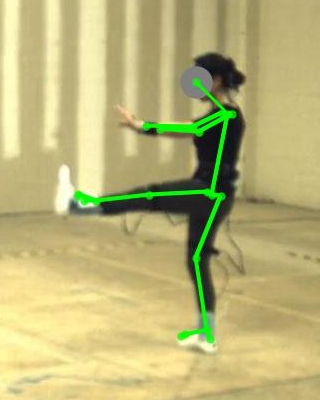} &
    	\includegraphics[width=.12\linewidth,valign=m]{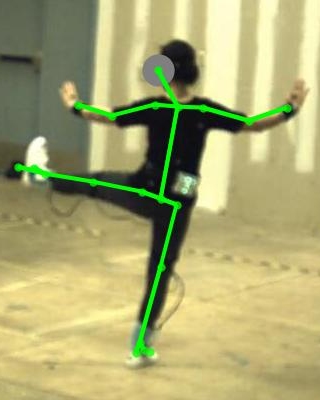} \\
    \end{tabular} \label{fig:PSU_TMM100:b}}
    \hfill
    \subfloat[]{\centering \includegraphics[width=0.25\linewidth,valign=m]{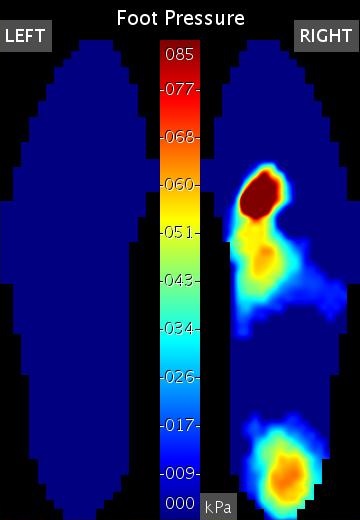} \label{fig:PSU_TMM100:c}}
    \vspace{-5pt}
    \caption{PSU-TMM100 data collection. \textbf{(a)}: Top-down view of motion capture environment. \textbf{(b)}: 3D pose computed from two video camera views. \textbf{(c)}: Foot pressure recorded synchronously using insole sensors.} 	\vspace{-15pt}
    \label{fig:PSU_TMM100}
\end{figure*}


    \subsection{Image-Based Dynamics Estimation}  
	Previous work in computer vision and graphics has explored estimation of ground contact forces from video and pose~\cite{brubaker2009,vondrakPhysicalSimulation,brubaker2010,Lv_2016,li2019motionforcesfromvideo}, but these estimates tend to be simple force vectors rather than the full foot pressure maps estimated in our work. 

	Using an RGB-D camera to record objects with known geometric and physical properties being manipulated by hands,~\cite{Pham2015} and~\cite{Pham2018} estimate the distribution of forces among the fingers in contact from vision-derived first and second order object kinematics
	using a network learned from hours of hand-object interactions.
This approach shows progress in estimating dynamics from video, but it is constrained to hands, requires sensing depth in addition to video, and is not ground truth validated. 

    \subsection{PSU-TMM100 Dataset} \label{sec:Dataset}
    Ground truth (GT) data for training and evaluation in this work is provided by simultaneously recorded video, motion capture, and foot pressure sensor data (Fig.~\ref{fig:PSU_TMM100}) from the PSU-TMM100 dataset~\cite{scott_2020}, where subjects perform approximately five-minute-long Taiji sequences by moving continuously through a set of complex body poses with a large range of joint articulations and limb orientations. PSU-TMM100 is the {\it only} available dataset that includes synchronized, sensor-measured recordings of these three modalities, making it a unique and valuable resource for learning to predict stability from imagery. The  dataset was collected using IRB-approved protocols (Study8085, initial approval 03-19-2018) with informed consent from all subjects. PSU-TMM100 demographics are ten subjects (five male and five female) with a wide range of experience performing Taiji (4--40~years, $\mu=13$, $\sigma=12$) and an average of ten performances (75,775--158,875~frames, $\mu=131,535$, $\sigma=26.749$) sampled at 50~Hz from each subject. As Taiji is a slow activity, all experiments use a sub-sampling of data to 5~Hz, reducing the computational resources needed for extensive training and testing on   5-minute motion sequences. Subjects have a broad range of mass (52.5--77.11~kg, $\mu=63.70$, $\sigma=6.95$) and height (1.54--1.80~m, $\mu=1.66$, $\sigma=0.08$). Four performances (takes) of Subject~2 (Takes 7, 9, 10, and 11) contain corrupt foot pressure data due to an insole sensor malfunction during recording. These outlier takes were discarded from the dataset prior to performing any evaluations reported in this paper.

        \subsubsection{Motion Capture} \label{Vicon_PiG}
        Ground Truth (GT) 3D pose in the dataset is provided by a Vicon motion capture system. Fig.~\ref{fig:pose_components:c} shows the 21 GT joints whose kinematics are generated by the Vicon Plug-in-Gait (PiG) model, which is based on the Conventional Gait Model (CGM)~\cite{Davis1991, Kadaba1990} originating from generic body segment inertial properties originally derived by Dempster from cadaver data~\cite{Dempster1955, Winter2009}.
        The PiG model also generates the GT CoM. A study comparing CoM location estimated by the Dempster parameters versus a more accurate reaction board method indicates a difference of 1\% or less expressed as a percentage of subject height (Fig 5 of \cite{Virmavirta2014}), which for this dataset is 16.8mm.

    	\subsubsection{Video Pose}
\begin{figure}[!t] \centering 
    \subfloat{\label{fig:pose_components:a}}
	\subfloat{\label{fig:pose_components:b}}
	\subfloat{\label{fig:pose_components:c}}
	\subfloat{\label{fig:pose_components:d}}
	\includegraphics[width=1\linewidth]{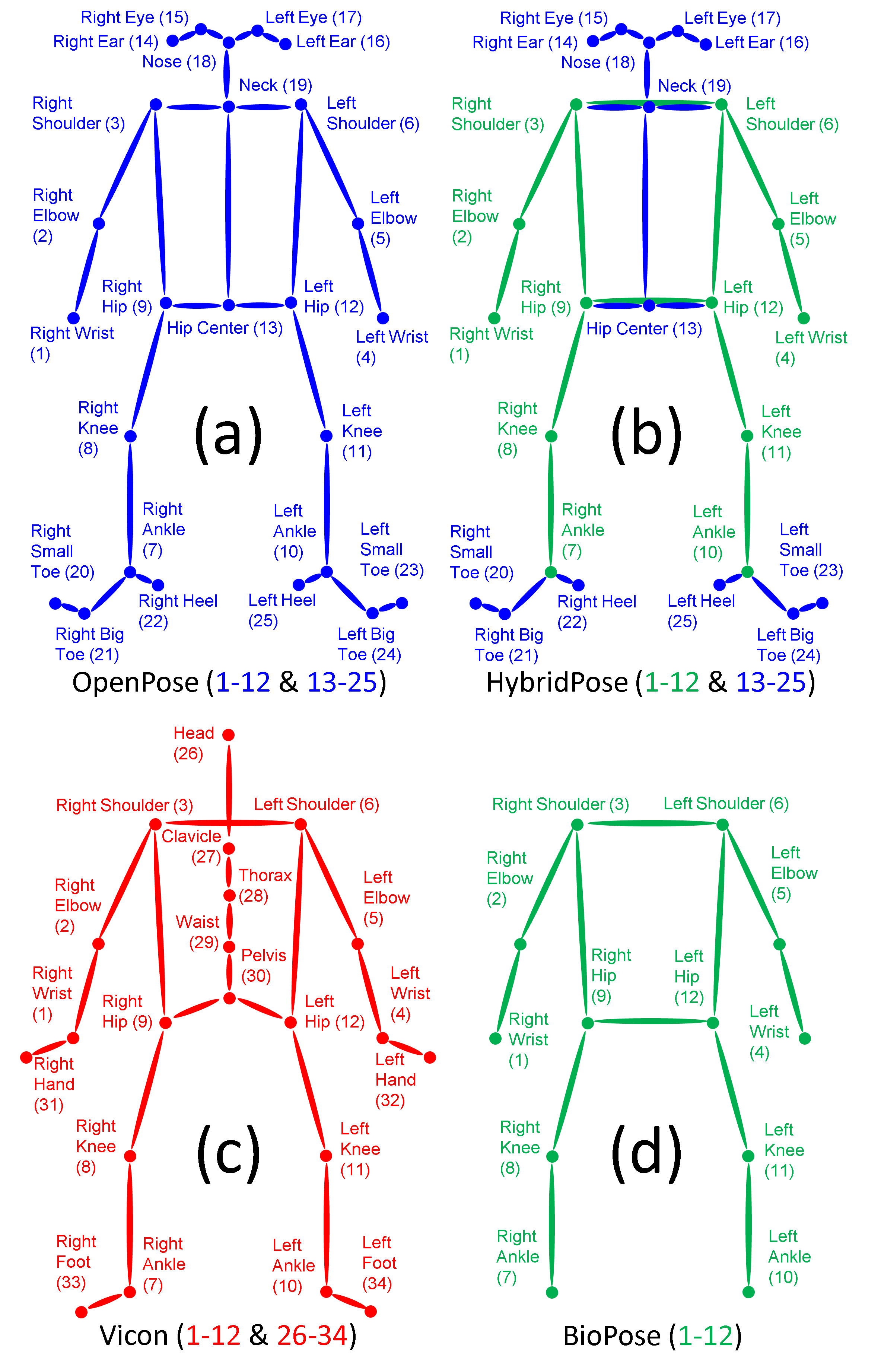}
	\vspace{-25pt}
	\caption{Comparison of body joints. \textbf{(a)}: OpenPose (OP)~\cite{wei2016convolutional,cao2017realtime}. \textbf{(b)}: HybridPose (HP) joints~$= BP \cup OP(13-25)$. \textbf{(c)}: ground truth from Vicon motion capture (GT). \textbf{(d)}: BioPose (BP)~$= GT \cap OP$. BP joints are common to all joint sets. HP is the 12 BP joints plus the remaining 13 OP joints.}
	\vspace{-15pt}
	\label{fig:pose_components}
\end{figure}

        Two HD video camera views spatiotemporally synchronized with the mocap system provide the data for estimating image-based pose (Fig.~\ref{fig:PSU_TMM100:b}). Four body joint configurations, OpenPose (OP), Mocap (GT), BioPose (BP), and HybridPose (HP), are used in this study (Fig.~\ref{fig:pose_components}). 
    	
    	We use OpenPose, an open-source 2D human pose estimator~\cite{wei2016convolutional,cao2017realtime} to predict 2D body joint locations, and two-view triangulation~\cite{hartley_2004} to reconstruct those 25 3D joint estimates (Fig.~\ref{fig:pose_components:a}). Triangulation of two views requires synchronized and calibrated cameras.  While estimation of 3D pose from a single camera view is desirable, state of art in that area is not yet mature, suffering from lower joint detection rates, decreased joint position accuracy, and inaccurate estimation of 3D body orientation with respect to gravity \cite{Kocabas_SPEC_2021}.

 There are 12 joints in common between GT (Fig.~\ref{fig:pose_components:c}) and OP, and we train the BioPose correction network from~\cite{scott_2020,Ravichandran_2020} to predict those 12 common joints (OP 1-12), improving their 3D biomechanical accuracy and generating BioPose (BP) joints (Fig.~\ref{fig:pose_components:d}). Lastly, HybridPose (HP) (Fig.~\ref{fig:pose_components:b}) is constructed by combining  BioPose joints (BP 1-12) with the 13 remaining non-overlapping OpenPose joints (OP 13-25). 
    	
		\subsubsection{Insole Pressure Measurement} \label{insole_HW}
        This research uses insole pressure data spatiotemporally synchronized with the video and motion capture data as the GT foot pressure (Fig.~\ref{fig:PSU_TMM100:c}) for training and testing the dynamics estimation networks. Insole sensors accurately measure foot pressure normal to the sensing plane, but with slower response times than force plates~\cite{chesnin_2000}, although still fast enough for human movement~\cite{Chevalier_2010}. 

\vspace{-5pt}
\section{Stability Components} \label{computing_stability}

	\subsection{Ground Truth (GT) CoM, BoS, and CoP }

		\subsubsection{Center of Mass (CoM)}
		The CoM is the 3D point about which the mass of a body is evenly distributed~\cite{beatty_2006}. The 3D CoM can be calculated for static and rigid objects, but the human body is much more complex with varying human tissue masses, body shape, and articulated body pose. Ground truth 3D CoM is calculated by Vicon PiG and is available directly from the motion capture portion of the dataset. Vicon PiG lower body model has been medically validated~\cite{carson_2001, stebbins_2006}, with~\cite{Baker2018} providing a thorough evaluation of PiG (and CGM) establishing its widespread use as well as the model's strengths and weaknesses. For purposes of this study, we treat PiG-modeled joints and calculated CoM as ground truth, following the precedent set by many biomechanical research laboratories and commercial applications~\cite{kainz_2017}. Specifically, the CoM is a 3D position; when projected onto the floor plane it is referred to as the 2D CoM.
	    
	    \subsubsection{Base of Support (BoS)}
		BoS is the convex hull that includes every point of contact that the subject makes with the supporting surface, including body parts (feet or hands) or support devices (crutches or walker)~\cite{BoS_2009}. Ground truth BoS is calculated from insole foot pressure maps after the feet are spatiotemporally localized using the mocap position of the ankles and toes to determine both location and orientation. This localized pressure map is used to create a binary mask of pressures above a minimum threshold (multiple thresholds are evaluated in Fig.~\ref{fig:PNS3_CoP} and~\ref{fig:PNS3_IoU}) from which a convex hull is calculated (Fig.~\ref{fig:composite_frames}).
		
		\subsubsection{Center of Pressure (CoP)}
		The CoP is the point at which the ground reaction force vector intercepts the supporting surface, calculated as the weighted sum of all forces acting between a physical object and its supporting surface~\cite{CoP_2009}. CoP is calculated as  a spatially weighted mean of all foot pressure samples in the XY plane of the floor  using the same localized pressure map used in the calculation of BoS (Fig.~\ref{fig:composite_frames}).

    \subsection{Image-Based CoM, BoS, and CoP}
    Input for our image-based CoM, CoP, and BoS computation begins with triangulated 3D poses calculated from two camera viewpoints (Fig.~\ref{fig:PSU_TMM100:b}). All experiments use the same Leave One subject Out (LOSO) data segmentation for cross-validation, ensuring that 
    the subject being evaluated has not been used in training. 

        \subsubsection{CoM Prediction} \label{CoM_prediction}
        We use a two-layer fully connected neural network called CoMNet to predict the CoM on a per-frame basis. CoMNet is trained to take 3D pose data and regress a 3D CoM location relative to the hip center. While CoMNet uses joint locations, it does not require the joint velocities/accelerations, subject measurements (height or weight), or the Dempster tables~\cite{Dempster1955} to predict a CoM. 
    
        CoMNet training is completed on a Nvidia Quadro K4000 with an RMSE loss function and an Adam optimizer. The network is empirically optimized to have 3072 wide fully connected input and hidden layers using batch normalization, a rectified linear unit, and 50~\% dropout regularization. CoMNet training takes approximately 2 hours for each of the 10 LOSO cross-validations. It takes 25 epochs with an initial learning rate of $5e-4$ and a piece-wise learning rate drop factor of 0.25 every 5 epochs. The CoMNet network and training weights will be available upon request following publication.

        \subsubsection{CoP and BoS Prediction}
        We use the PressNet-Simple 3D (PNS3) network from~\cite{scott_2020} for image-based foot pressure predictions. While OpenPose joint data are shown in~\cite{scott_2020} to be the best input for predicting foot pressure, we evaluate motion capture, HybridPose, and OpenPose data for foot localization for all takes of PSU-TMM100. The calculation of CoP and BoS follows the same calculation steps as the ground truth process but replaces sensor inputs with image-based data. CoP and BoS are used for comparing each image-based configuration against ground truth motion capture and insole pressure data. CoP accuracy is evaluated using Euclidean distance between predicted and ground truth locations. BoS is evaluated using the Intersection over Union (IoU) metric, also known as the Jaccard Index~\cite{jaccard_1901}.

\vspace{-5pt}
\section{Stability Metrics} \label{selected_metrics}
After a thorough review of the human balance and stability literature, e.g.,~\cite{Bruijn_2013,jian_1993,Lugade_2011,Hof_2005,Slobounov_1997,Haddad_2006}, two well-established stability metrics were selected for evaluation in this paper: CoMtoCoP and CoMtoBoS (Table~\ref{tbl:stability_metrics}). These two metrics can be calculated from the available data modalities and are well suited for a non-repetitive performance like Taiji  that focuses on maintaining biomechanical stability. Both metrics are easily understandable and collectively use all three stability components CoP, CoM, and BoS.  A more extensive set of experiments that include additional stability metrics xCoMtoBoS, CoMvtoBoS, and TTC (time to contact) can be found in the first author's Ph.D.~thesis \cite{scottThesis2022}.

\begin{figure}[!t] \centering
    \includegraphics[width=.9\linewidth]{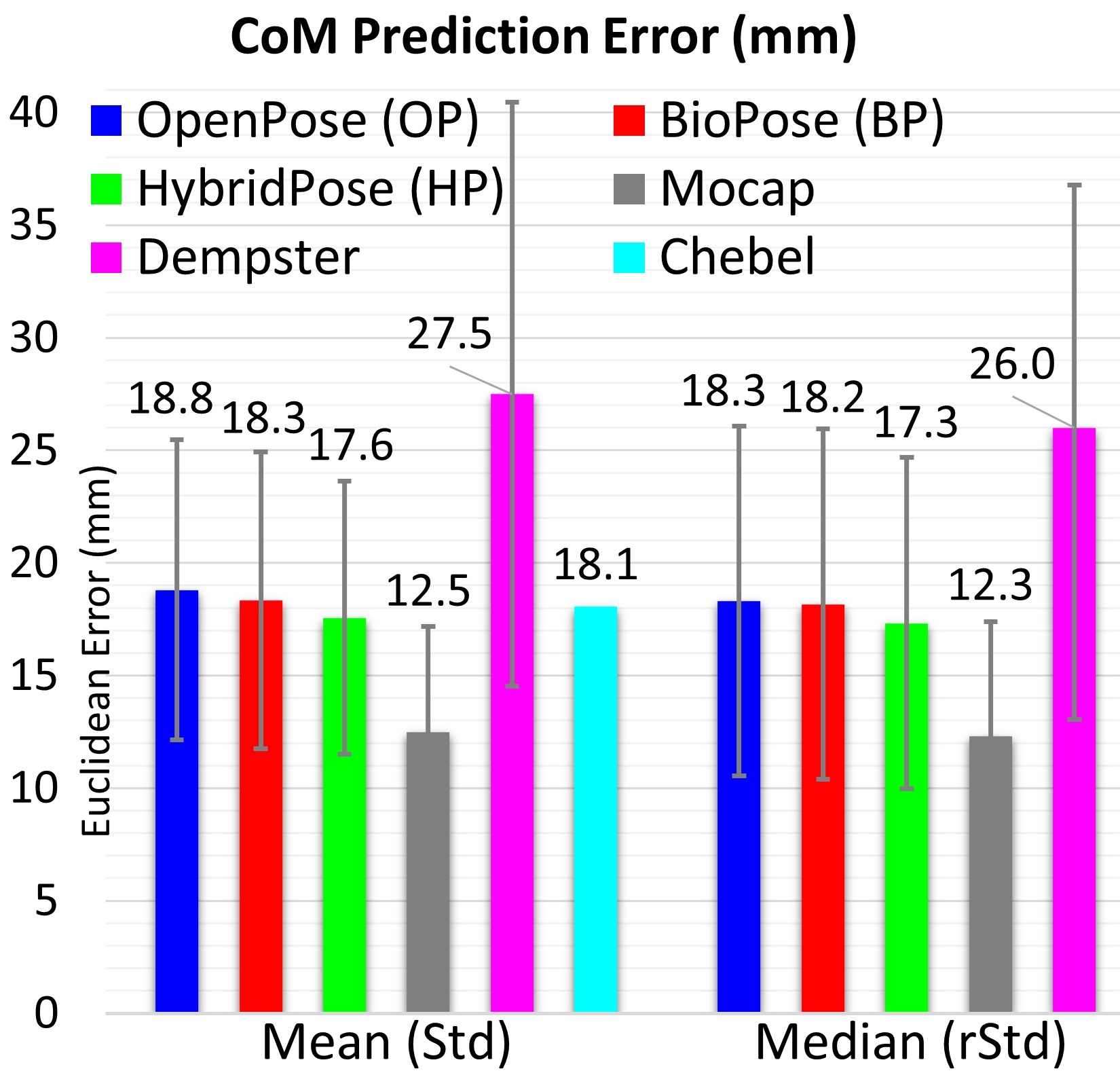}
    \vspace{-10pt}
    \caption{3D CoM prediction error (mm) of CoMNet when input pose comes from: \textcolor{blue}{OP}, \textcolor{red}{BP}, \textcolor{green}{HP} (best), and \textcolor{gray}{Mocap} (practical limit) as well as \textcolor{magenta}{Dempster}~\cite{Dempster1955} applied to HP joints and \textcolor{cyan}{Chebel}~\textit{et al.}~\cite{chebel_2021}. Statistics provided: mean~(Std) and median~(rStd) as compared to GT CoM derived from Vicon PiG (Section~\ref{Vicon_PiG}). Results are based on poses when all body joints are detected. Robust standard deviation (rStd)~=~1.4826 times median absolute deviation (MAD)\cite{madestimator}.} 
    \label{fig:CoM_error} \vspace{-15pt}
\end{figure}


\begin{figure*}[!t] \centering
    \subfloat{\centering \includegraphics[width=.3\linewidth]{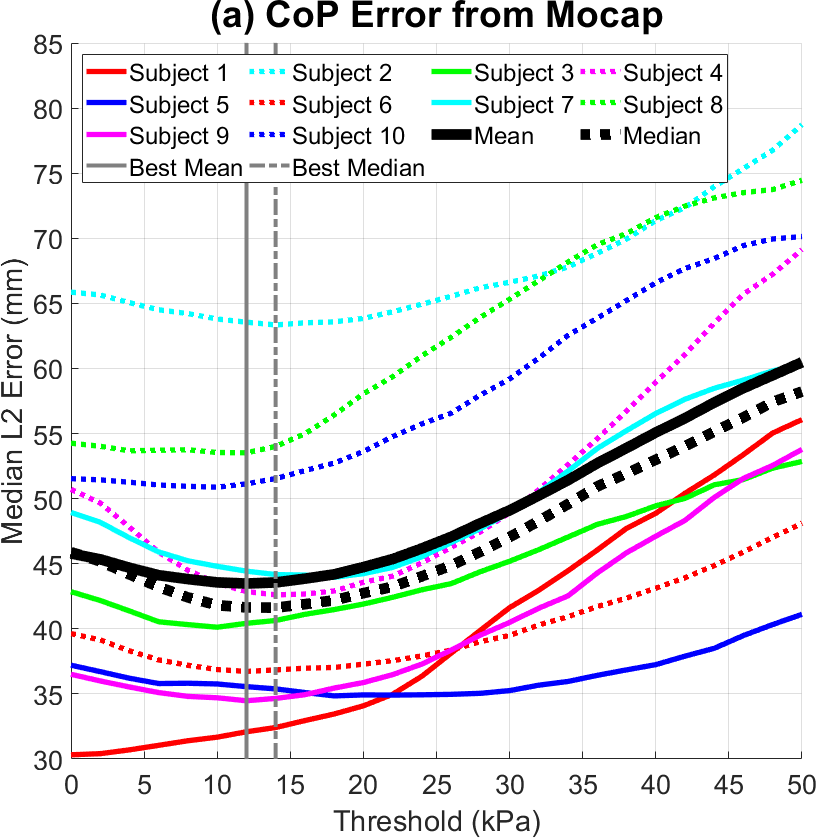} \label{fig:PNS3_CoP:a}} \hfill
    \subfloat{\centering \includegraphics[width=.3\linewidth]{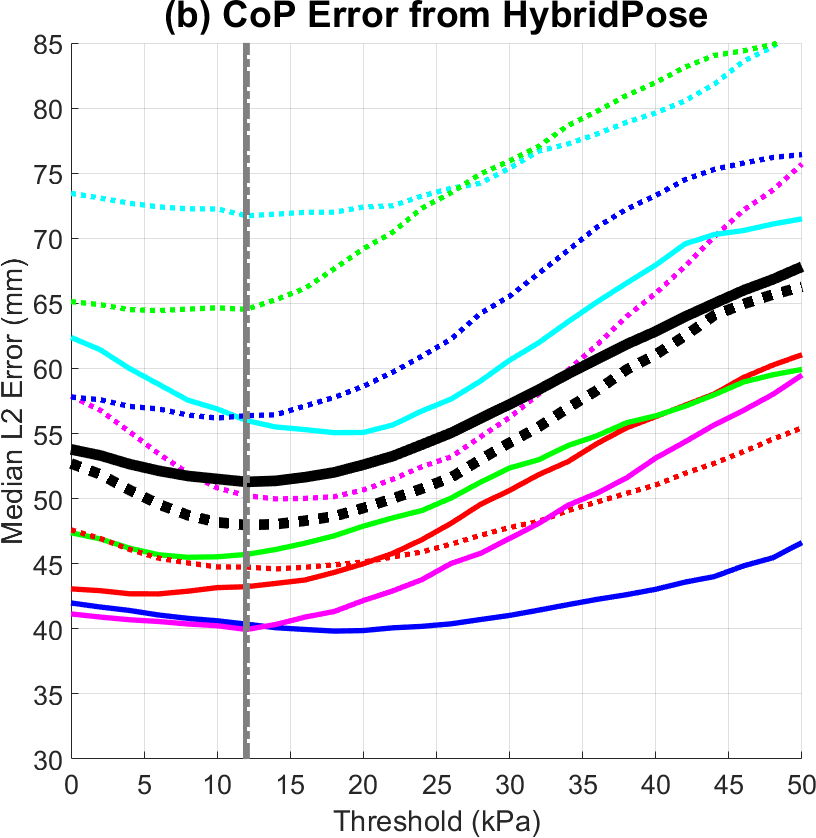} \label{fig:PNS3_CoP:b}} \hfill
    \subfloat{\centering \includegraphics[width=.3\linewidth]{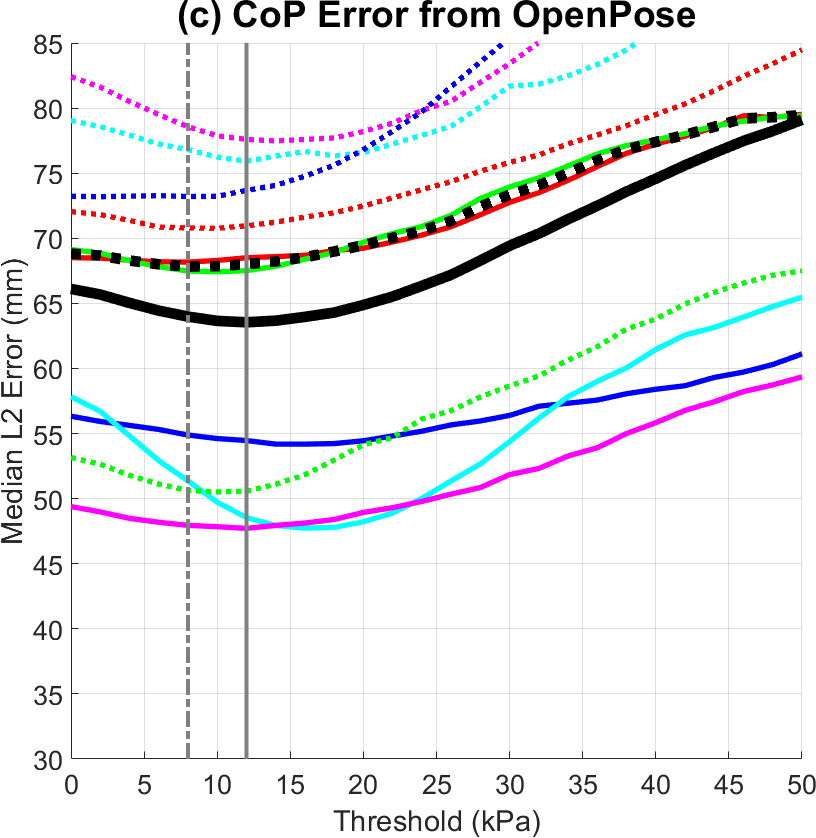} \label{fig:PNS3_CoP:c}}
    \vspace{-10pt}
    \caption{CoP $\ell^2$ error (mm) relative to sensor-based GT (lower better). All results use PNS3~\cite{scott_2020} predicted pressure distribution maps and foot localization from \textbf{(a)} Mocap, \textbf{(b)} HybridPose, or \textbf{(c)} OpenPose, respectively. BioPose localization is excluded due to a lack of required joint locations, toes and heels (Fig.~\ref{fig:pose_components:d}). HybridPose input (b) provides the best image-based result. The x-axis shows increasing thresholds (kPa) where pressures below the threshold are set to zero.} 
    \label{fig:PNS3_CoP}
    \vspace{-15pt}
\end{figure*}


\begin{figure*}[!t] \centering
    \subfloat{\centering \includegraphics[width=.3\linewidth]{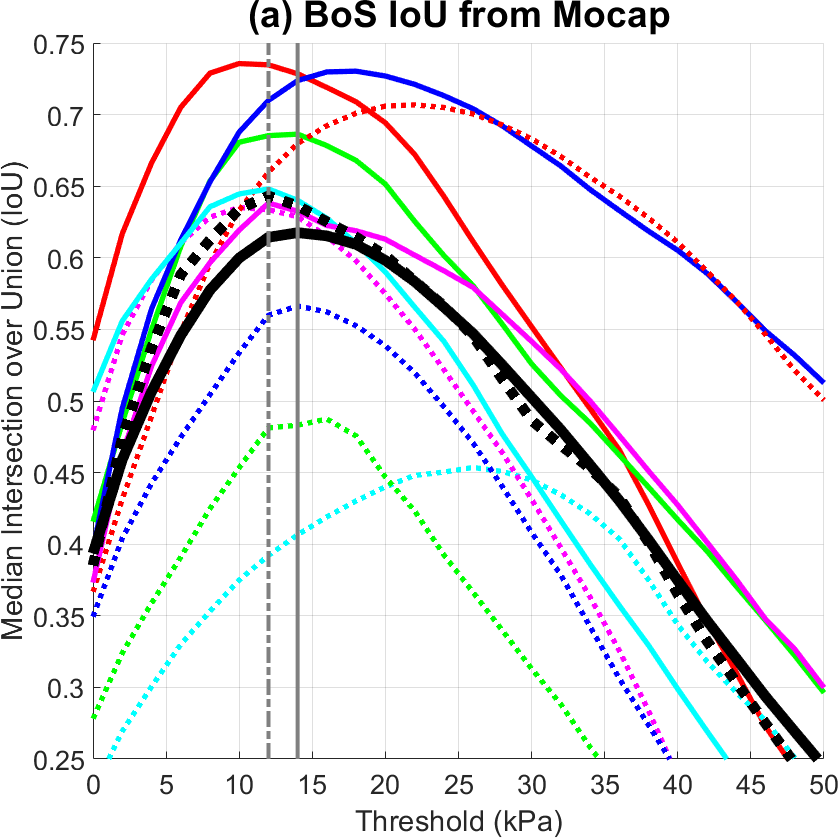} \label{fig:PNS3_IoU:a}} \hfill
    \subfloat{\centering \includegraphics[width=.3\linewidth]{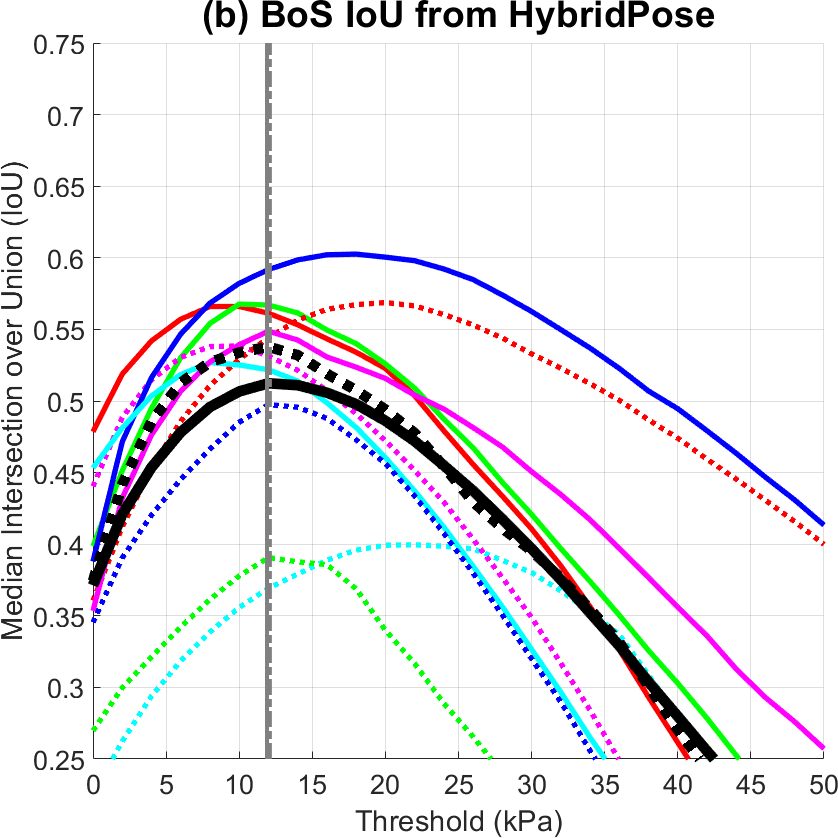} \label{fig:PNS3_IoU:b}} \hfill
    \subfloat{\centering \includegraphics[width=.3\linewidth]{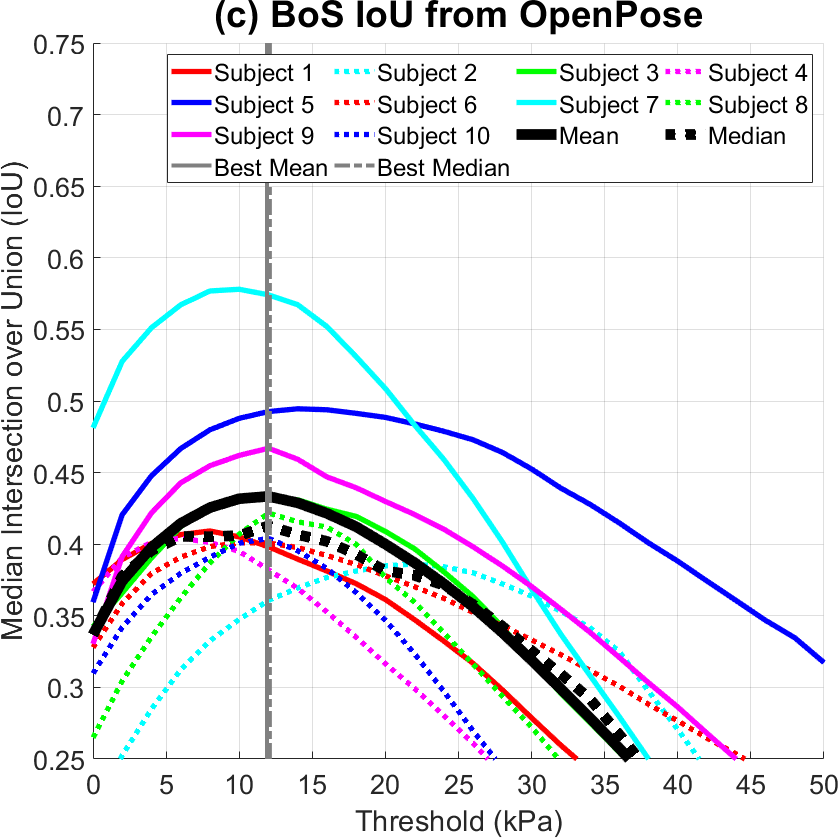} \label{fig:PNS3_IoU:c}}
    \vspace{-10pt}
    \caption{BoS accuracy using IoU relative to sensor-based GT (higher better). All results use PNS3~\cite{scott_2020} predicted pressure distribution maps and foot localization from \textbf{(a)} Mocap, \textbf{(b)} HybridPose, or \textbf{(c)} OpenPose, respectively. BioPose localization is excluded due to a lack of required joint locations, toes and heels (Fig.~\ref{fig:pose_components:d}). HybridPose input (b) provides the best image-based result. The x-axis shows increasing thresholds (kPa) where pressures below the threshold are set to zero.} 
    \vspace{-10pt}
    \label{fig:PNS3_IoU}
\end{figure*}

    \subsection{CoMtoCoP}
    The Euclidean distance between a subject's 2D CoM and CoP measures the spatial difference between ground reaction force and gravitational force  (Table~\ref{tbl:stability_metrics}, Equation~\ref{eq:CoMtoCoP}) ~\cite{jian_1993,Chaudhry_2011}. Conceptually, the further apart these two points are, the greater the potential for instability~\cite{Hof_2005}. While keeping the two points close together may seem advantageous, in dynamic tasks trained athletes can tolerate greater excursion compared to those not trained~\cite{Ambegaonkar_2013}, as can the young compared with the old~\cite{Cavanaugh_1999}. Therefore, subjects who are better at maintaining their stability (perceptually and physically) can allow this distance to become large while still being able to avoid instability. CoMtoCoP is a nonnegative distance measurement typically reported in millimeters, with values normally near zero. A larger variance during a performance indicates subjects with better stability control.

    \subsection{CoMtoBoS}
    The Euclidean distance from the 2D CoM to the border of the BoS quantifies both the magnitude and condition of mechanical imbalance (Table~\ref{tbl:stability_metrics}, Equation~\ref{eq:CoMtoBoS})~\cite{Lugade_2011}. CoMtoBoS magnitude is the distance from the CoM to the nearest point on the BoS boundary; CoMtoBoS is positive if CoM is inside the BoS and negative otherwise. 
    Negative values indicate imbalance/instability that requires intervention to prevent an eventual fall while positive values indicate mechanical balance and stability~\cite{Lugade_2011}. There is an inherent maximum positive distance and no limit in the negative direction, but small positive values indicate better stability control~\cite{Hof_2005}. 

    
\vspace{-5pt}
\section{Results} \label{results}
\vspace{-5pt}
	\subsection{CoM Prediction}
	
	Fig.~\ref{fig:CoM_error} evaluates CoM location estimates produced by various configurations of CoMNet against Vicon PiG CoM estimates (GT) provided with the dataset.  HybridPose CoMNet, that is, CoMNet trained to take HybridPose 3D joint estimates as input, outperforms CoMNet trained on either BioPose or OpenPose joints.  
    HybridPose CoMNet is thus the best performing variant using purely image-based inputs, with mean (+/- std) location error of 17.6 (6.1)~mm.  
    Noting that GT CoM locations provided by PiG are computed by a segmental method using Vicon Mocap joints and Dempster table parameters, two additional baseline methods are evaluated. "Dempster" is the Dempster segmental method applied to image-estimated  HybridPose 3D joints.  The larger mean error of 27.5 (13.0)~mm indicates that CoMNet is compensating for differences between 3D joints estimated by HybridPose and Mocap.  A second baseline, "Mocap", is our ComNet trained using GT Mocap joint data as input. The mean error of 12.5~mm establishes a practical limit on CoMNet accuracy when input joints are as accurate as possible.

    ComNet HybridPose outperforms BioPose input joints, indicating that useful information is learned by CoMNet when the additional 13 OpenPose joints are combined with BioPose joints. Additionally, all image-based configurations produce similar and consistent results. Using only image-based pose input, CoMNet establishes a state-of-the-art better than the mean Euclidean error of 18.1~mm achieved by Chebel~\textit{et al.}~\cite{chebel_2021} that requires subject measurements and inertial sensors. 
	
     

	\subsection{CoP}\label{CoP_results}
	Fig.~\ref{fig:PNS3_CoP} shows results of the PNS3 network architecture~\cite{scott_2020} on all valid performances in the dataset for overall mean/median (black solid/dashed) and per-subject mean (colors) accuracy. We compare ground truth foot localization with HybridPose and OpenPose localization to quantify the performance of image-based localization. All three foot localization plots show peak performance between 10~kPa~\cite{Keijsers_2009} and 15~kPa~\cite{Hsiao_2002} (indicated by gray vertical lines), which are  commonly used threshold and peak accuracies, respectively. There are three key observations:
	\begin{enumerate}
    	\item HybridPose localization provides the best fully image-based CoP results due to improved ankle accuracy from the BioPose network, with 51.3/48.0~mm (mean/median) error being a small increase from the mocap localization error of 43.5/41.6~mm.
    	\item HybridPose does not uniformly improve CoP results over OpenPose as Subjects 7 and 8 (light and dark blue plots in Fig.~\ref{fig:PNS3_CoP}) are better with OpenPose.
    	\item The all-performances results are similar to the one-take-per-subject results reported in~\cite{scott_2020}.
    \end{enumerate}
    
	\subsection{BoS}\label{BoS_results}
	BoS was also evaluated on all valid performances to determine how different foot localization methods affect IoU accuracy (Fig.~\ref{fig:PNS3_IoU}). Foot localization accuracy affects IoU of BoS more than CoP error as foot pressure pixels of small magnitude can cause large changes in the size and shape of the BoS while having little change on CoP.
	There are three key observations:
	\begin{enumerate}
    	\item HybridPose localization provides the best (higher is better) image-based IoU results (improved ankle accuracy from BioPose network) with 51.24/53.78\% (mean/median); a small decrease from the mocap localization IoU of 61.76/64.32\%.
    	\item HybridPose does not uniformly improve IoU results over OpenPose as Subjects 7 and 8 (light and dark blue plots in Fig.~\ref{fig:PNS3_IoU}) are better with OpenPose.
    	\item The all-performances results of each subject are similar to the one-take-per-subject results in~\cite{scott_2020}, suggesting those one-take results were statistically representative.
	\end{enumerate}

	\subsection{Stability Metrics} \label{metric_results}
	To evaluate image-based estimation of stability metrics,
CoMtoCoP and CoMtoBoS are calculated from combinations
of ground truth (\textcolor{red}{GT}) and image-based estimates (\textcolor{blue}{IM})
over three data channels (foot pressure, foot localization and CoM)
for 8 combinations total.
The image-based estimates used are: 1) PNS3 with OpenPose for foot pressure (shown in~\cite{scott_2020} to be
the state-of-the-art); 2) 
HybridPose for foot localization (shown in Fig.~\ref{fig:PNS3_CoP:b} and~\ref{fig:PNS3_IoU:b} to produce the best CoP and BoS results); and 3) HybridPose CoMNet for CoM (shown in 
Fig.~\ref{fig:PNS3_IoU:b} to provide the most
accurate CoM estimate).


\begin{table*}[!t] \centering
    \caption{Combinatorial study of correlation coefficient (r-value) with Mean Absolute Error (MAE) and standard deviation (Std) of distance from GT calculations for both CoMtoCoP and CoMtoBoS compared to All Ground Truth in mm. CoP and BoS are directly computed by combining pressure and localization. Input combination order: foot pressure - foot localization - center of mass. Data sources are ground truth (\textcolor{red}{GT}) or image-based predictions (\textcolor{blue}{IM}). Values are the mean for all 10 LOSO experiments. Key combinations are \colorbox{red!100}{All Ground Truth}, \colorbox{blue!100}{\textcolor{white}{Only GT Foot Pressure}}, and \colorbox{green!100}{Fully Image-based} corresponding to Fig.~\ref{fig:stability_metric_combos}. Only complete performances are included and all results are p~<=~0.001 except p~<~0.05(*) and p~>~0.05(+).}
    \vspace{-5pt}
    \resizebox{1\textwidth}{!}{%
    \begin{tabular}{|c|cccccccc|}\hline
    \multicolumn{9}{|c|}{Combinatorial Study of Ground Truth and Image-based Inputs using r-value (Std) \& MAE (Std) in mm} \\ \hline 
    Pressure-Location-CoM & \textcolor{red}{GT}-\textcolor{red}{GT}-\textcolor{red}{GT} & \textcolor{red}{GT}-\textcolor{red}{GT}-\textcolor{blue}{IM} & \textcolor{red}{GT}-\textcolor{blue}{IM}-\textcolor{red}{GT} & \textcolor{red}{GT}-\textcolor{blue}{IM}-\textcolor{blue}{IM} & \textcolor{blue}{IM}-\textcolor{red}{GT}-\textcolor{red}{GT} & \textcolor{blue}{IM}-\textcolor{red}{GT}-\textcolor{blue}{IM} & \textcolor{blue}{IM}-\textcolor{blue}{IM}-\textcolor{red}{GT} & \textcolor{blue}{IM}-\textcolor{blue}{IM}-\textcolor{blue}{IM} \\ \hline \rowcolor{Gray}
    \multicolumn{9}{|c|}{CoMtoCoP} \\ \hline
    r-value (Std) & \cellcolor{red!100}{1.00 (0.00)} & 0.88 (0.20)	& 0.88 (0.08) & \cellcolor{blue!100}{\textcolor{white}{0.79 (0.18)}} & 0.39 (0.09) & 0.34 (0.13) & 0.35 (0.09) & \cellcolor{green!100}{0.31 (0.10)} \\ \hline
    MAE (Std) & \cellcolor{red!100}{0.00 (0.00)} & 10.14 (23.75) &  15.90 (22.88) & \cellcolor{blue!100}{\textcolor{white}{18.12 (34.70)}} & 37.37 (51.47) & 40.00 (61.21) & 40.28 (54.77) & \cellcolor{green!100}{41.92 (62.81)} \\ \hline \rowcolor{Gray}
    \multicolumn{9}{|c|}{CoMtoBoS} \\ \hline
    r-value (Std) & \cellcolor{red!100}{1.00 (0.00)} & 0.83 (0.24)	& 0.86 (0.08) & \cellcolor{blue!100}{\textcolor{white}{0.75 (0.21)}} & 0.32 (0.15)* & 0.25 (0.14)+ & 0.27 (0.12)* & \cellcolor{green!100}{0.22 (0.12)*} \\ \hline
    MAE (Std) & \cellcolor{red!100}{0.00 (0.00)} & 9.12 (22.11) & 11.82 (16.98) & \cellcolor{blue!100}{\textcolor{white}{14.76 (28.95)}} & 25.47 (35.35) & 28.40 (45.10) & 28.95 (38.54) &  \cellcolor{green!100}{31.07 (46.44)} \\ 
    \hline \rowcolor{Gray}		
    \hline
    \end{tabular}
    }
    \vspace{-5pt}
    \label{tab:pr_evaluation} \label{tab:stability_error} 
\end{table*}

\begin{figure*}[!t] \centering
    \includegraphics[width=.95\linewidth]{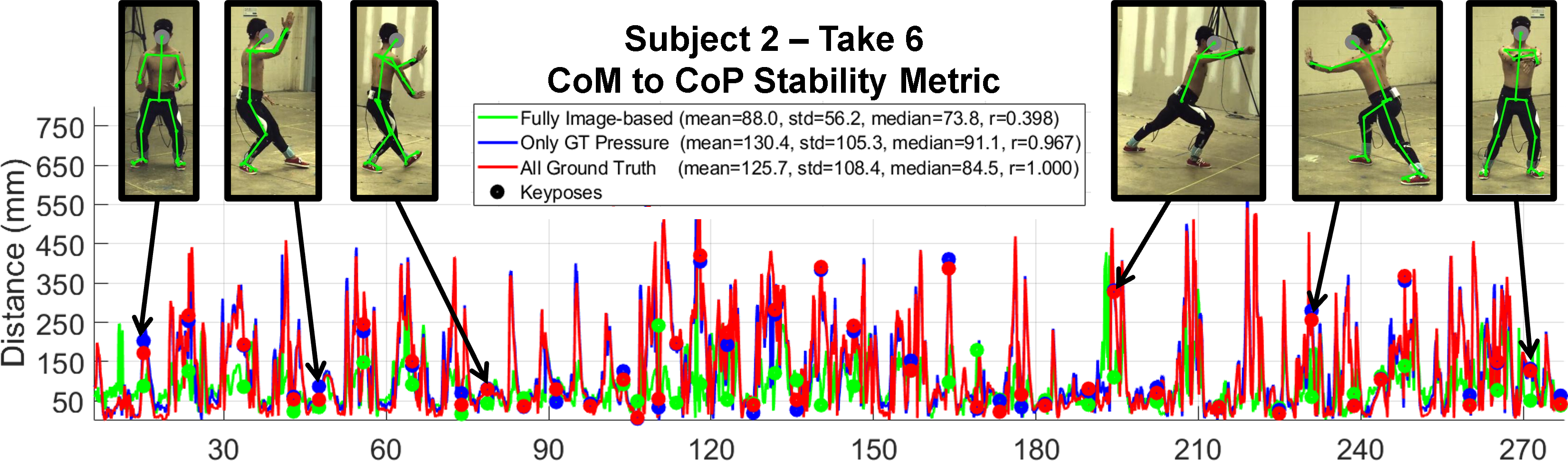} \\ \vspace{5pt}
    \includegraphics[width=.95\linewidth]{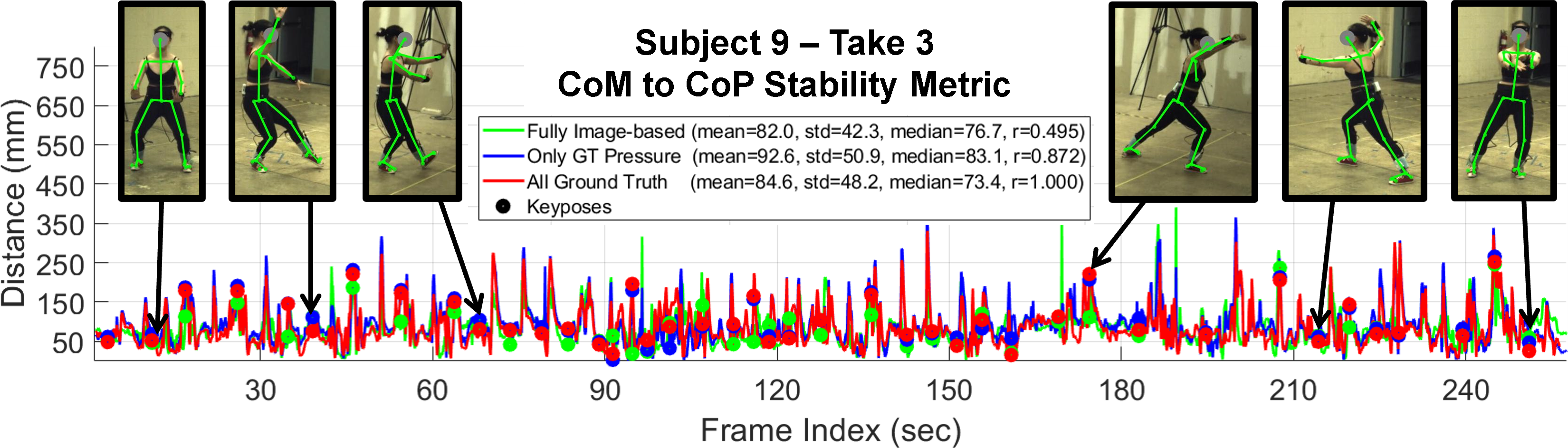} \\ \vspace{-10pt}
    \caption{Examples of CoMtoCoP results highlighting similar trends of all three combinations presented: fully ground truth (\textcolor{red}{red}), ground truth foot pressure with all other inputs image-based (\textcolor{blue}{blue}), and fully image-based (\textcolor{green}{green}). Based on CoMtoCoP r-value: Subject2 - Take6 (top) is the best for Only GT Pressure and Subject9 - Take3 (bottom) is the best for Fully Image-based. Plots include image call-outs of key poses with video joint overlay, mean, standard deviation, median, and r-value for each combination. Plot colors are related to highlighted columns of the comprehensive results in Table~\ref{tab:stability_error}. The red line heavily occludes blue and green because of very strong correlation.} \vspace{-10pt}
    \label{fig:stability_metric_combos}
\end{figure*}

\begin{figure*}[!t] \centering
    \renewcommand{\tabcolsep}{1.5pt}
    \resizebox{1\textwidth}{!}{%
    \begin{tabular}{c c}
    \includegraphics[width=.5\linewidth]{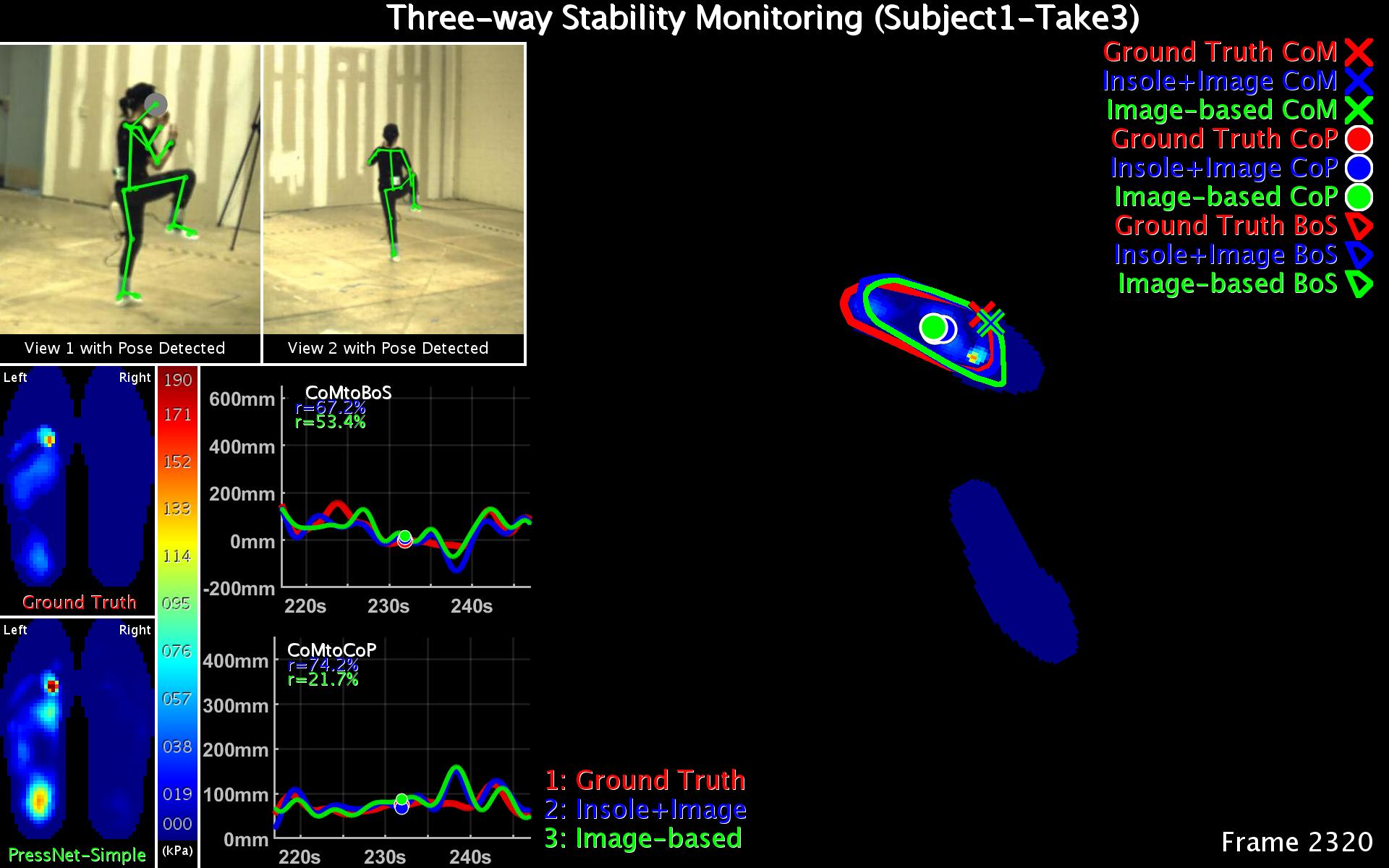} &
    \includegraphics[width=.5\linewidth]{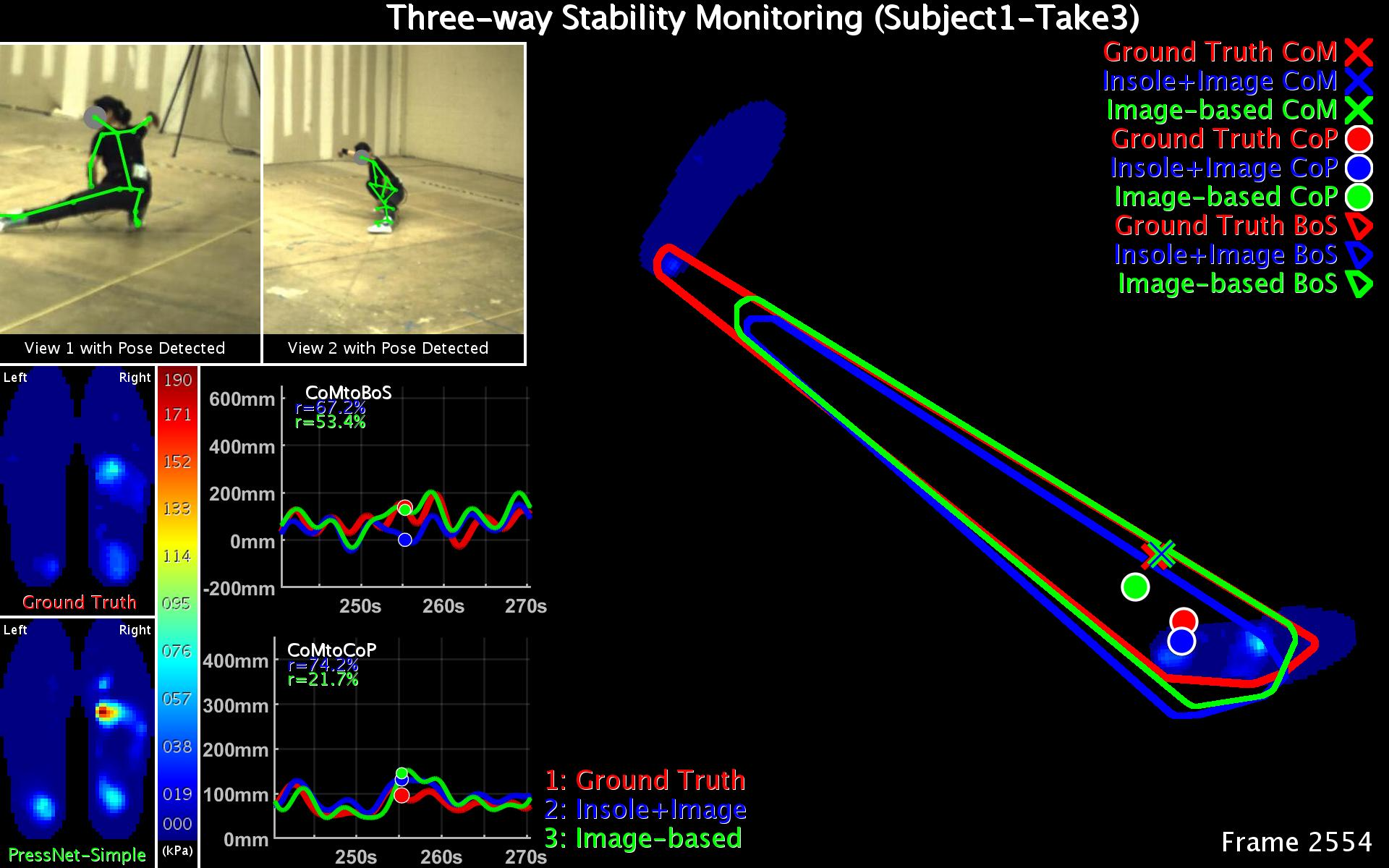} \\
    \includegraphics[width=.5\linewidth]{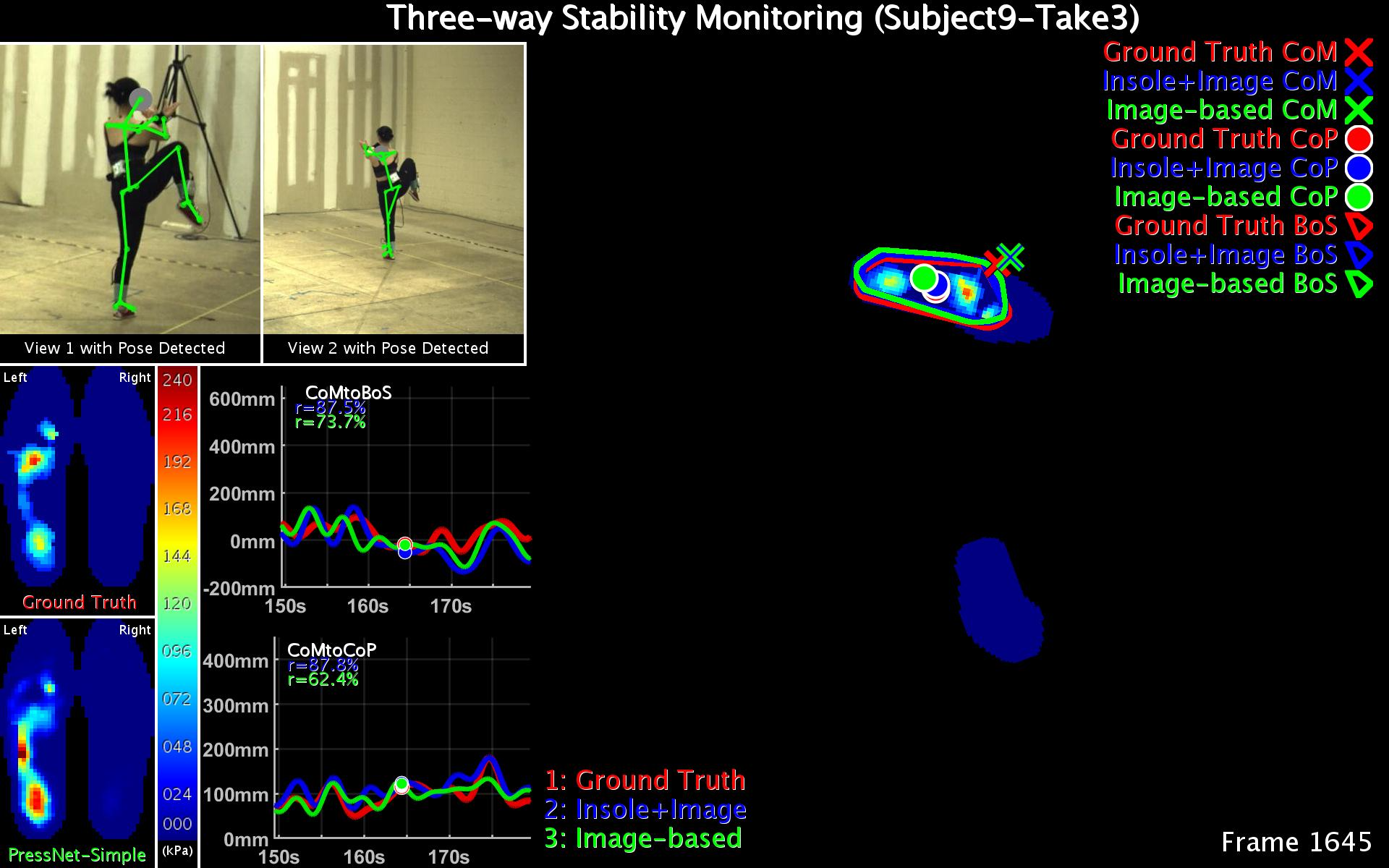} &
    \includegraphics[width=.5\linewidth]{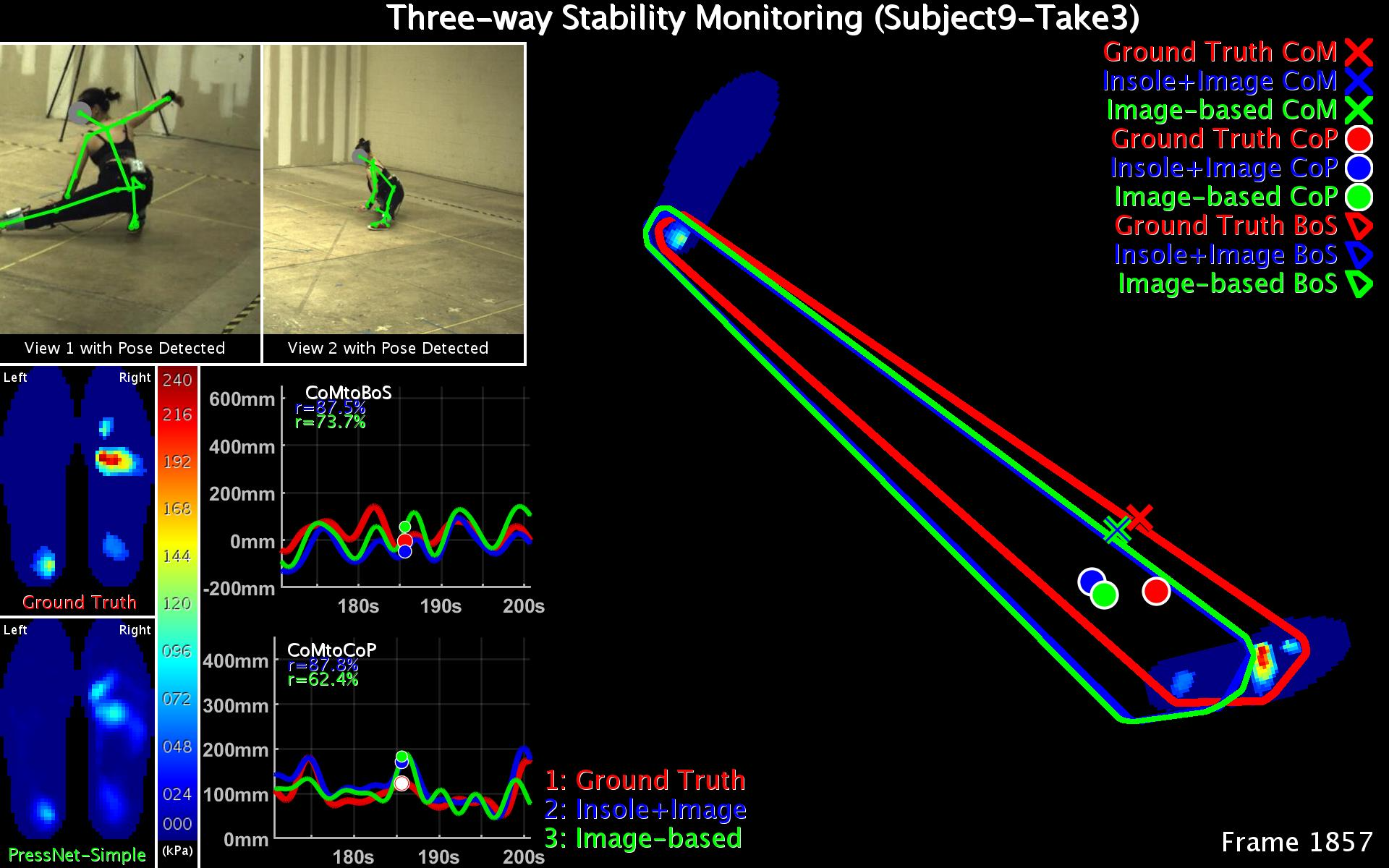} \\
    \end{tabular} 
    } 
    \vspace{-10pt}
    \caption{Image-based stability results compared to GT of Subject 1 - Take 3 (Top) and Subject 9 - Take 3 (Bottom) representing the best performance (based on r-value) for CoMtoCoP and CoMtoBoS, respectively. Each frame includes input images, motion capture, and foot pressure plus output stability components and metrics. Samples include single foot leg lift (left) and double foot lunge (right) poses selected from full performances. Results color map: {\textcolor{red}{Ground Truth (red)}}, {\textcolor{blue}{Image+Insole (blue)}}, and {\textcolor{green}{Image-based (green)}}.} 
    \label{fig:composite_frames} \vspace{-5pt}
\end{figure*}

Stability metric values computed from each combination are compared to fully ground truth estimates to determine correlation coefficient (r-value) and statistical significance (p-value). The mean and std of r-values across all 10 Leave One Subject Out (LOSO) experiments are reported in Table~\ref{tab:pr_evaluation}.
Using sensor-based pressure measurements (\textcolor{red}{GT}) with image-based inputs (\textcolor{blue}{IM}) for localization and CoM (Table~\ref{tab:pr_evaluation} \colorbox{blue!100}{\textcolor{white}{blue}}) produces correlation r-values of 0.79 and 0.75 for both CoMtoCoP and CoMtoBoS, respectively. Using image-based localization and CoM eliminates the need for motion capture hardware. Switching to image-based foot pressure, i.e., fully image-based stability (Table~\ref{tab:pr_evaluation} \colorbox{green!100}{green}), yields reduced but still positive correlation coefficients of 0.31 and 0.22, respectively.

    From Table~\ref{tab:pr_evaluation} results it is observed that image-based foot pressure prediction has the largest effect on the r-values with 1.00 to 0.39 and 1.00 to 0.32 decreases relative to GT input for CoMtoCoP and CoMtoBoS, respectively. Conversely, image-based foot localization effects on r-values are relatively small with 1.00 to 0.88 and 1.00 to 0.86 decreases, respectively, while image-based CoM effects are also small with 1.00 to 0.88 and 1.00 to 0.83 decreases, respectively. These results indicate that image-based foot pressure estimation has the largest room for improvement at approximately five times the r-value effect of image-based localization or CoM estimation.



    Table~\ref{tab:stability_error} also reports the Mean Absolute Error (MAE) for each of the 8 combinations of stability estimates relative to GT.  MAE consistently increases when the stability metrics use more image-based input data, while standard deviation increases primarily when image-based foot pressure is included. The only minor difference between the two metrics is that CoMtoBoS has both lower r-values and MAE across most combinations when compared to CoMtoCoP. Since both BoS and CoP derive from foot pressure, it is expected that both metrics would have generally similar behavior (Fig.~\ref{fig:PNS3_CoP} and~\ref{fig:PNS3_IoU}). Additionally, the overall lower values are expected since CoMtoBoS has a data range that includes negative values, unlike CoMtoCoP which has to be $\geq 0$.

    Fig.~\ref{fig:stability_metric_combos} visualizes  CoMtoCoP stability metric results for two performances by plotting a fully ground truth result computed using motion capture and insole sensors compared to two combinations that use some or all image-based input data: 1) image-based localization and CoM prediction with GT foot pressure, which eliminates the need for motion capture sensor requirements and 2) fully image-based predictions that eliminate both motion capture and foot pressure sensor requirements. As compared to the red curve showing (\colorbox{red!100}{All Ground Truth}) results, Subject 2 - Take 6 represents the best r-value results for the blue curve (\colorbox{blue!100}{\textcolor{white}{Only GT Foot Pressure}}) while Subject 9 - Take 3 represents the best r-value results for the green curve (\colorbox{green!100}{Fully Image-based}). Each plot includes six keypose images with detected joint overlay. Both plots show strong overlap between the blue and red curves due to their strong correlation (r=0.97 and 0.87), while the green curves exhibit only partial overlap with the red ground truth curves, reflecting only moderate correlation (r=0.40 and 0.50).

    Fig.~\ref{fig:composite_frames} focuses on the qualitative results of calculating imaged-based stability components (CoP, BoS, and CoM) and stability metrics (CoMtoCoP and CoMtoBoS). The frames show two Taiji poses from performances by Subject 1 - Take 3 (top) and Subject 9 - Take 3 (bottom), representing the best r-value results (0.48 and 0.50, respectively) when using\colorbox{green!100}{Fully Image-based} estimation with PNS3 (OpenPose) foot pressure prediction, HybridPose for foot localization, and CoMNet from HybridPose for CoM prediction (green in Table~\ref{tab:stability_error}). These information-rich frames (Fig.~\ref{fig:composite_frames}) facilitate at a glance a qualitative comparison of estimated components CoM, CoP, BoS and stability measures CoMtoCoP and CoMtoBoS computed from either all ground truth values (red), using ground truth insole pressure but otherwise image-based estimates (blue), or fully image-based estimation (green).
    
    There are two key takeaways from this analysis.  First, a fully image-based approach (eliminating the need for foot pressure sensors and motion capture) produces stability estimates that are positively correlated with GT (CoMtoCoP r~=~0.31 p~<~0.001, CoMtoBoS r~=~0.22 p~<~0.043).  Second,  a hybrid approach using insole foot pressure sensor data combined with image-based foot localization and CoM prediction (eliminating need for motion capture hardware) produces stability estimates that are strongly correlated with GT estimates (CoMtoCoP r~=~0.79 p~<~0.001, CoMtoBoS r~=~0.75 p~<~0.001).

	\subsection{Computational Costs}
    For each sampled time instance, all data processing and analysis are performed in under 2 seconds using an 8 core PC with 64~GB of RAM, without optimizing for speed of processing. Of this time, over 1 second is used to estimate the 3D image-based pose while the remaining time is used for foot pressure and CoM estimation combined with stability calculations.

\begin{figure*}[!t] \centering
    \includegraphics[width=.49\linewidth]{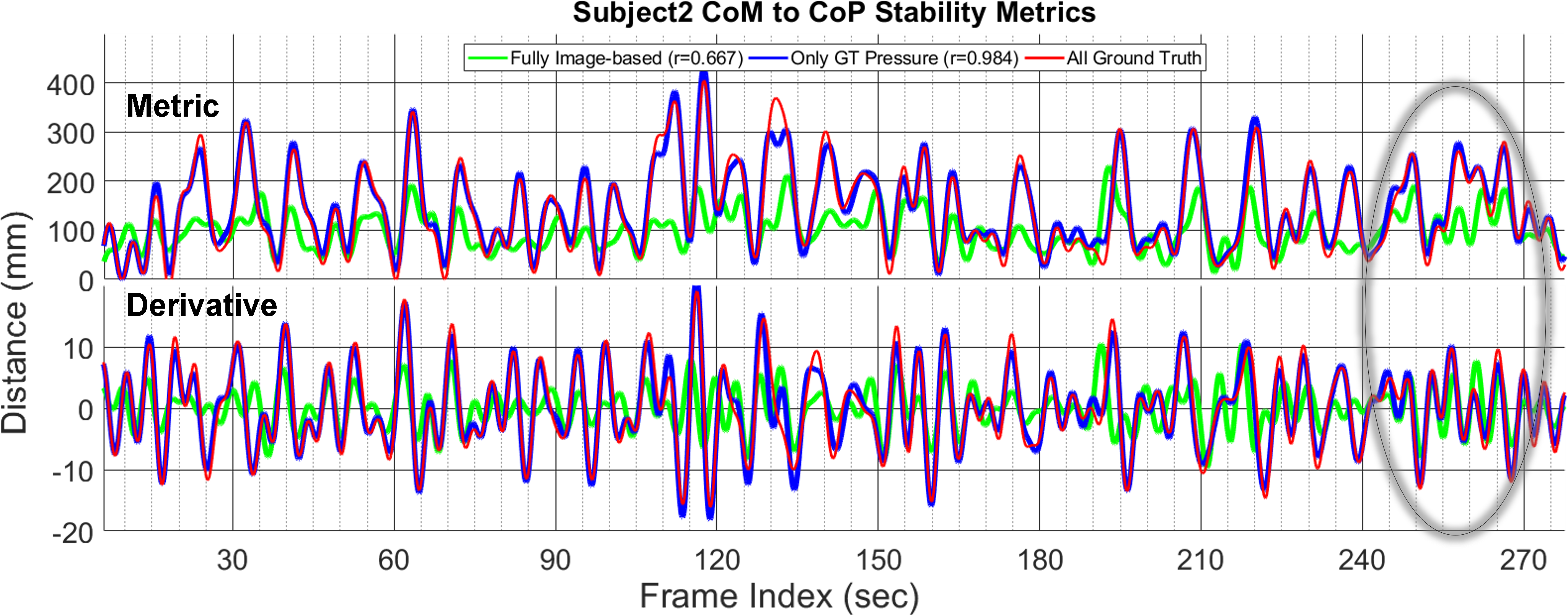} 
    \includegraphics[width=.49\linewidth]{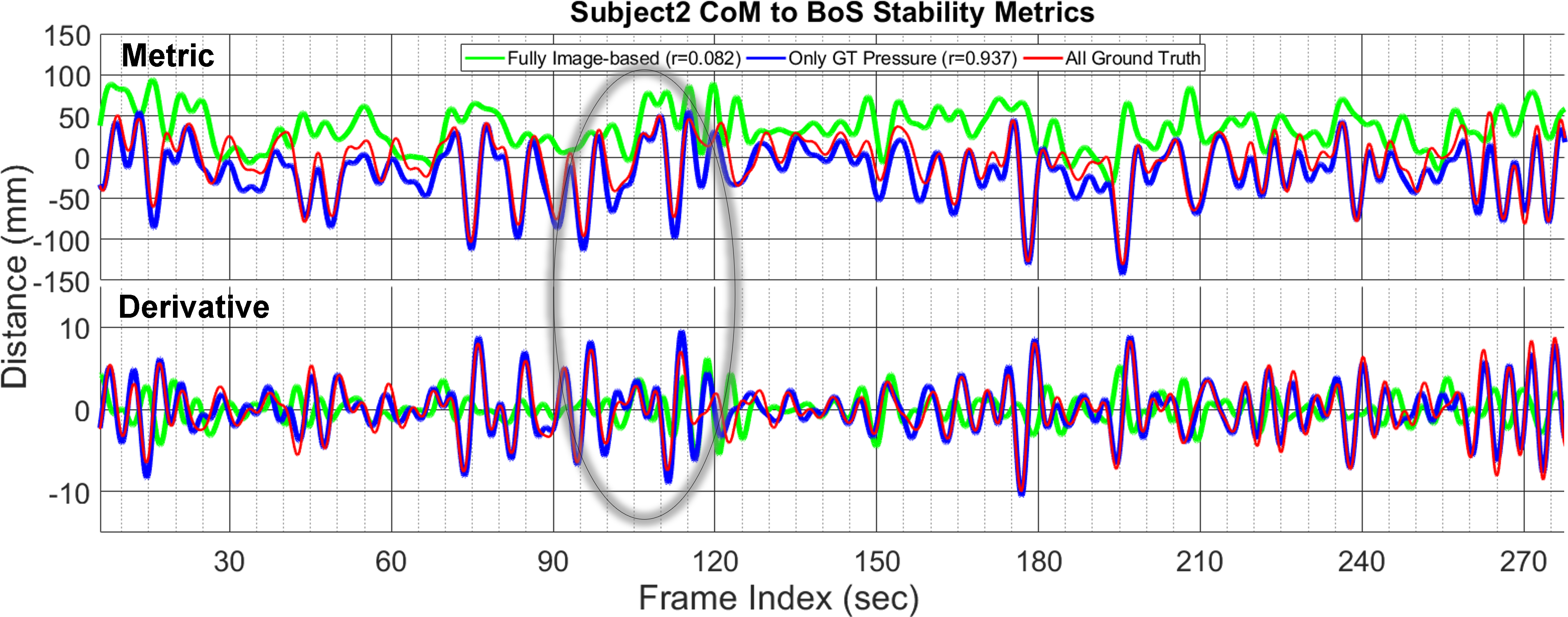} \\ \vspace{5pt}
    \includegraphics[width=.49\linewidth]{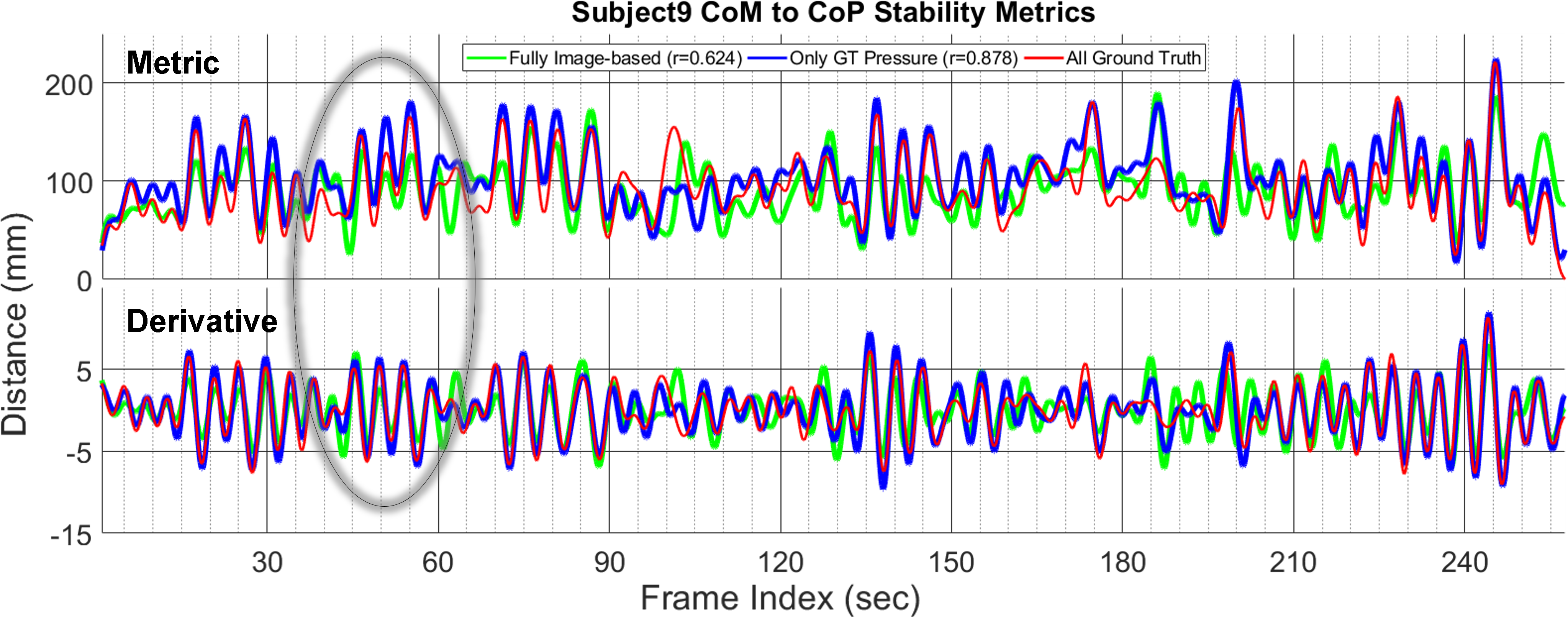} 
    \includegraphics[width=.49\linewidth]{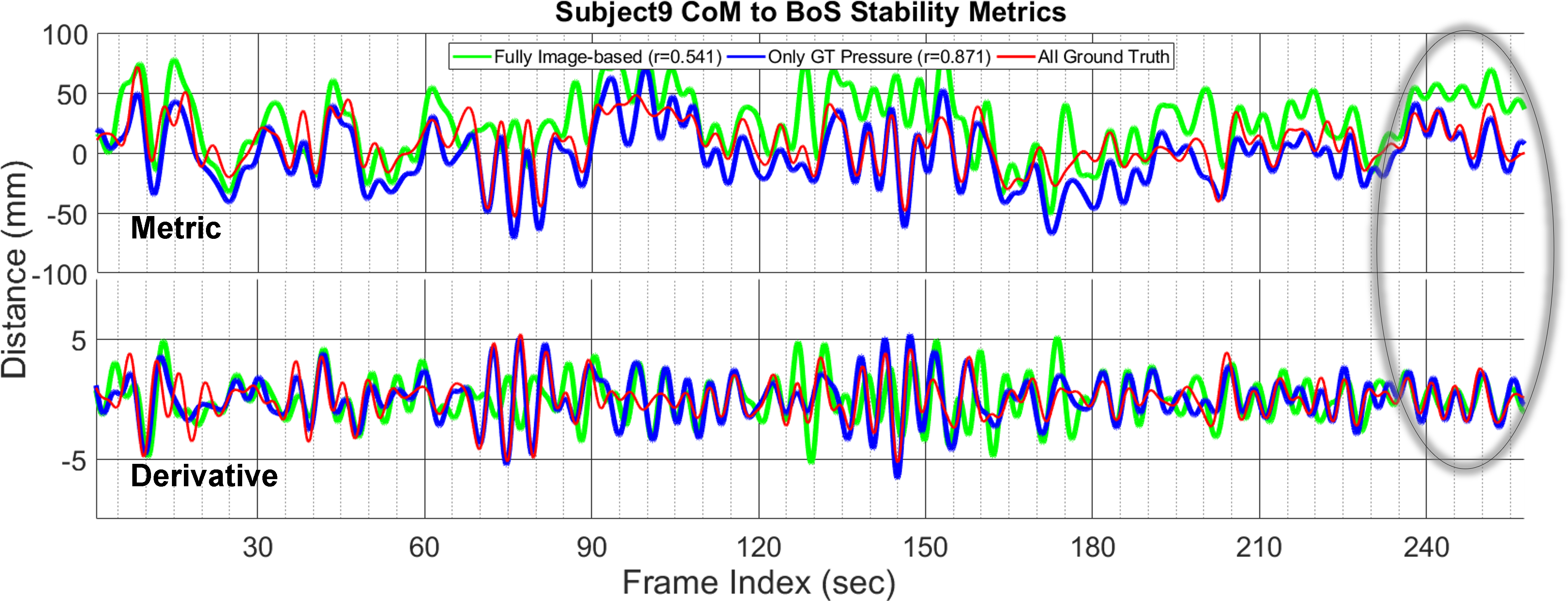}
    \vspace{-5pt}
    \caption{Smoothed CoMtoCoP (left) and CoMtoBoS (right) stability and their derivative curves from Fig.~\ref{fig:stability_metric_combos} highlighting similar trends between ground truth (\textcolor{red}{red}), image-based with insole sensor only (\textcolor{blue}{blue}) of Subject 2 (top), and fully image-based (\textcolor{green}{green}) of Subject 9 (bottom). Gray circles highlight when derivatives are well correlated.} \vspace{-10pt}
    \label{fig:stability_metric_CoMtoCoP}
\end{figure*}

    \subsection{Stability Trends Analysis}
    Based on the stability analysis completed at 5~Hz sampling rate (Section~\ref{sec:Dataset}), low frequency content is modeled using a zero-lag, low-pass filter (0.2~Hz). Fig.~\ref{fig:stability_metric_combos} stability metric data are low-pass filtered to generate low frequency stability trends (Fig.~\ref{fig:stability_metric_CoMtoCoP}).
    The computed ``Only GT Pressure'' and ``Fully Image-based'' curves for Subject 2 (r-values of 0.98 and 0.67 compared to low-pass filtered ground truth) and Subject 9 (r-values of 0.88 and 0.62) illustrate similar trends; i.e., upward/downward-sloping curves, indicating relative consistency with GT stability measures. 

\vspace{-5pt}
\section{Conclusion} \label{Conclusion}
This work demonstrates that image-based stability quantification is computationally feasible (Fig.~\ref{fig:composite_frames}). Using 2D pose extracted from two RGB cameras, 3D pose is triangulated and used to predict foot pressure and to compute CoM. The predicted foot pressure is further combined with image-based foot localization to calculate BoS and CoP. These three stability components (CoM, CoP, and BoS) are combined to calculate image-based predictions of stability metrics CoMtoCoP and CoMtoBoS, which are quantitatively shown in Section~\ref{metric_results} to have significant positive correlation with the GT stability metric values. Stability metrics computed from image-based pose combined with pressure sensors produce strong correlations of 0.79 and 0.75, respectively. 
Fully image-based stability estimates have lower yet  
positive correlations of 0.31 and 0.22 with ground truth, respectively, indicating the potential feasibility for fully image-based stability estimation upon with improvements to image-based foot pressure prediction.

CoMNet predicts image-based 3D CoM from 3D pose with a mean Euclidean error of 17.56~mm, outperforming the state-of-the-art method using body-worn inertial sensors~\cite{chebel_2021}, and predicting an error nearly as low as the expected error in ground truth motion capture calculations~\cite{Virmavirta2014} while using only image-based data. Additionally, the work originally published in~\cite{scott_2020} reporting CoP and BoS results for one-take-per-subject sub-sampling is validated here for all valid dataset performances (Fig.~\ref{fig:PNS3_CoP} and \ref{fig:PNS3_IoU}), confirming the sub-sampling in~\cite{scott_2020} was a representative cross-section.

Computing quantified stability measures exclusively from imagery substantially reduces the need for expensive, physically encumbering equipment that constrains data collection to laboratory environments. The presented methods therefore may enable smart health interventions in real-world conditions based on timely image-based evaluation of human stability.

\vspace{-5pt}
\section*{Acknowledgment}
This research is supported in part by NSF grant IIS-1218729 and the Penn State College of Engineering Dean's office.



\bibliographystyle{IEEEtran}
\bibliography{Biblio-Database.bib}
\end{document}